\documentclass[11pt]{article}
\usepackage{acl}
\usepackage{times}
\usepackage{latexsym}
\usepackage[T1]{fontenc}
\usepackage[utf8]{inputenc}
\usepackage{microtype}
\usepackage{inconsolata}
\usepackage{graphicx}
\usepackage{amsmath}
\usepackage{enumitem}
\usepackage{booktabs}
\usepackage{xcolor}
\usepackage{array}
\usepackage{multirow}
\usepackage{makecell}
\usepackage{titlesec}
\titlespacing*{\section}{0pt}{10pt plus 1pt minus 2pt}{4pt plus 0pt minus 1pt}
\titlespacing*{\subsection}{0pt}{8pt plus 1pt minus 2pt}{3pt plus 0pt minus 1pt}
\usepackage{balance}
\usepackage{amssymb}

\definecolor{gaingreen}{RGB}{22, 101, 52}
\definecolor{lossred}{RGB}{159, 18, 57}

\title{Not All Speech Is Intent:\\
Adaptive Self-Correcting Inference Layer for Post-ASR False Wake-Up}

\author{
  Preeti Saraswat$^{\dagger}$ \quad
  Divya Neelagiri$^{*}$ \quad
  Anil Yadav$^{*}$ \\
  Samsung Research America, Mountain View, CA, USA \\
  \texttt{\{p.saraswat, d.neelagiri, Anil.y1\}@samsung.com}
  \thanks{$^{*}$Equal contribution.
  $^{\dagger}$Corresponding author.}
}
\begin{document}
\maketitle

\begin{abstract}
False wake-up activations remain a persistent challenge in conversational AI.
Speech phonetically similar to a device's wake word can produce a
syntactically valid and semantically coherent ASR transcript that the assistant
incorrectly executes. Most existing systems make a single intent decision in isolation, without a mechanism to learn from recurring errors over time or adapt to individual users through personalized learning.
We introduce the \textit{Feedback-Driven Adaptive Self-Correcting Inference
Layer (ASCIL)}, a complementary post-ASR correction framework that re-evaluates
wake-up intent before response generation by fusing acoustic embeddings,
linguistic cues, device context, and patterns from past misclassifications. ASCIL interprets implicit
signals, including hesitation, disengagement, and silence, and explicit signals,
including cancellation and repetition, as automatically inferred, noisy
behavioral indicators of potential misclassification. These signals drive online
pattern updates without manual annotation, whereas the intentional/unintentional
reference labels used for offline evaluation are human-annotated. It generalizes from
prior errors, applies corrective adjustments at inference time, and continuously
updates in parallel with natural-language execution. Evaluated on a proprietary
dataset of 3{,}667 interactions with human-annotated intentional/unintentional
reference labels spanning 14 acoustic and contextual conditions, ASCIL achieves
\textbf{54.27\% relative error reduction} on a session-disjoint subset
constructed from baseline failures, and up to \textbf{24.39\% relative error
reduction} at threshold 0.90 on the issue-tagged evaluation slice. These gains
are achieved while improving intentional acceptance rates, with a median added
latency below 60\,ms in the reported benchmark.
\end{abstract}

\section{Introduction}
\label{sec:intro}

Consider a user watching television when the dialogue contains a phrase
phonetically similar to the device's wake word. The ASR (Automatic speech recognition) system
transcribes the output cleanly in a complete valid transcript.
Nothing in the transcript reveals that the activation was unintentional.
The assistant responds. The user ignores it, or says ``stop.'' The same
thing happens the next day. Nothing changes.

This scenario exposes a structural gap in deployed conversational AI. While prior work has addressed
same-turn device-directedness classification and other methods, \textbf{a closed feedback loop that learns from user
reactions after a wake-up has already been handled remains largely unaddressed}. Once a wake-up
event produces a valid transcript, execution happens.
There is no mechanism to ask whether the activation was truly intended,
and no mechanism to remember that this pattern caused an error before.
Two compounding problems make this worse.

First, \emph{static classifiers}: production systems are trained
centrally on population-level data and cannot adapt to individual users.
A user who has a regional accent, and keeps the
television on while cooking experiences a systematically different error
distribution than the user who works in a noisy environment, and existing systems offer no
mechanism to correct for either pattern individually.

Second, \emph{no post-ASR feedback loop}: user reactions to false
activations like cancellations, repetitions, disengagement are not involved in current systems improvement or learning. These signals are a source of noisy
behavioral supervision that the system could learn from. The user may provide an implicit or explicit behavioral signal after a potentially misclassified interaction, and the system can use that signal as noisy supervision for future pattern updates.

We introduce a \textbf{Feedback-Driven ASCIL} that
addresses both problems. Operating post-wakeup and post-ASR, the system
fuses multimodal features to revalidate wake-up intent before response
generation. When an automatically inferred feedback signal indicates a
potential misclassification, the system extracts a generalized pattern capturing the acoustic, linguistic, and
contextual conditions of that error, stores it compactly on-device, and
applies it at future inference to proactively suppress similar
misclassifications. The base ASR and NLU (Natural Language Understanding) models are not modified or retrained. Our contributions are:
\begin{itemize}[leftmargin=*,itemsep=2pt,topsep=4pt]
 \item A \emph{post-ASR feedback loop} that is complementary to existing pre-ASR and
  same-turn device-directedness classifiers: such a classifier can sit as the initial
  call in our pipeline, with our feedback loop improving its input conditions over time.

  \item A \emph{feedback-derived supervision mechanism} that interprets implicit
  signals (hesitation, disengagement, silence) and explicit signals
  (cancellation, repetition) as automatically inferred, noisy indicators of
  potential misclassification. These signals support online pattern updates
  without additional manual annotation and are distinct from the
  human-annotated reference labels used for offline evaluation.

  \item An \emph{ASCIL} that extracts generalized patterns from prior
  feedback-derived error hypotheses, applies them at inference, and updates
  continuously in parallel with NL execution, with a median added
  correction-path latency below 60\,ms in the reported benchmark.

  \item Empirical results on a real-world proprietary dataset showing
  \textbf{54.27\% relative error reduction} on held-out failing cases and
  up to \textbf{24.39\%} on the issue-tagged evaluation slice at
  operating thresholds.
\end{itemize}

\section{Related Work}
\label{sec:related}

Much prior work emphasizes wake-word detection, false-trigger mitigation, or same-turn device-directedness classification, with comparatively limited attention to feedback-driven post-decision adaptation from subsequent user reactions. ~\citet {Schonherr2020} systematically characterized accidental triggers
across smart speakers. ~\citet{chen2021fakewake}
investigated the FakeWake phenomenon, generating fuzzy words that fool
wake-word detectors and proposing detector-level remedies.~\citet{wu2020falsewake} proposed
monophone-based background modeling for two-stage on-device detection.
~\citet{agarwal2020ftm} explored complementary language
modeling to reduce false triggers during decoding, and ~\citet{rudovic2023ftm} unified multiple invocation types in a
multi-task architecture. ~\citet{lin2025vad} addressed domain
mismatch through personalized VAD models.~\citet{baljekar2014wordspotting} and ~\citet{chaudhary2023wordspotting} addressed keyword spotting
efficiency on-device.

Personalization for ASR was explored by 
~\citet{McGrawpersonalized} using quantized LSTM acoustic models, and
SpecAugment~\citet{park2019specaugment} provided robust augmentation for
training. ~\citet{gasic2017online} demonstrated on-line reward learning for
dialogue policy; ~\citet{mazumdar2024dialogue} surveyed
continual learning in dialogue systems.

A closely related industrial disclosure is
~\citet{kanungo2026system}, which describes post-ASR false wake-up
suppression using textual, acoustic, and contextual evidence together with
adaptive decision thresholds. Our work differs in that ASCIL focuses on
post-interaction behavioral feedback as noisy supervision and stores
generalized user-specific error patterns for future inference, without
retraining the base ASR or NLU models. Other Closest research work is ~\citet{mallidi2018device} which addressed post-ASR device-directedness
classification, determining whether an utterance was directed at the device using signals
available immediately after transcription. Their approach and ASCIL are complementary. A same-turn device-directedness classifier can provide the initial decision, while ASCIL
uses subsequent user behavior to refine future decisions.

To our knowledge, prior false-wake and device-directedness systems have not
reported a mechanism that converts automatically inferred post-interaction
feedback into persistent, generalized, on-device corrective patterns without
retraining the base models. ASCIL differs in three ways: it learns from user
reactions after a wake-up has been executed or suppressed, maintains
user-specific memory of recurring misclassification patterns, and applies those
patterns at inference time without updating the base ASR or NLU parameters.

\section{System}
\label{sec:system}

\subsection{Overview}

\begin{figure*}[t]
  \centering
  \includegraphics[width=\textwidth]{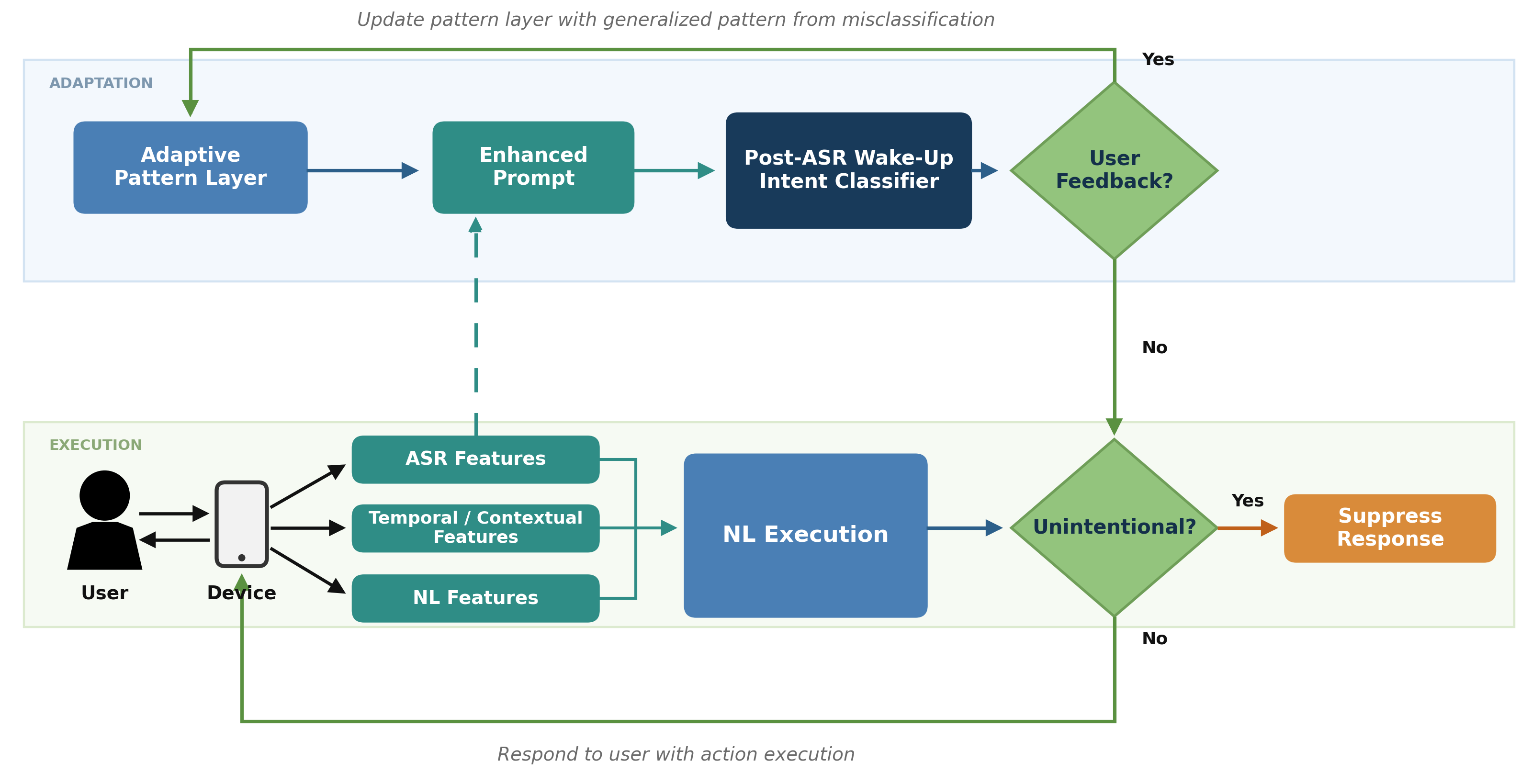}
  \caption{The adaptive correction pipeline. Post-ASR multimodal
  features feed a base post-ASR wake-up intent classifier. User feedback signals trigger
  pattern extraction. Stored patterns are injected
  as context at future inference, proactively correcting misclassifications
  without modifying base ASR or NLU models.}
  \label{fig:architecture}
\end{figure*}

Figure~\ref{fig:architecture} illustrates the pipeline. At each
interaction step $t$, the system receives a multimodal feature vector
$X_t \in \mathbb{R}^d$ consolidating four signal categories: acoustic
descriptors (pitch, SNR, speech rate, HNR, barge-in events, pre-inverse-text-normalization (pre-ITN) ASR text), linguistic
cues from ASR and NLU (token and sentence embeddings, POS distributions,
named entity presence, syntactic completeness, topic relevance),
temporal and device context (session history, foreground application,
wake modality, device motion), and adaptive patterns from prior
misclassifications. 
A base classifier maps $X_t$ to a binary intent
prediction $\hat{Y}_t \in \{0, 1\}$, where $1$ denotes an unintentional
activation. Let $Y_t$ denote the human-annotated intentional/unintentional
reference label, used only for offline evaluation, and let $F_t$ denote an
automatically inferred behavioral feedback signal, used only for online pattern
updates. ASCIL then refines the prediction:
\[
\hat{Y}'_t = C\!\left(X_t,\, P_u^{(t)}\right)
\]
where $P_u^{(t)}$ is the on-device user-specific pattern memory at
time $t$, and $C$ is the correction function.

\subsection{Post-ASR Intent Classification}

The base classifier operates in two modes. For standard activations, it
predicts intentional vs.\ unintentional from $X_t$. When the input
contains a user feedback signal like user cancellation, a repeated command,
or sustained silence following an unexpected response, the classifier
does not produce an intent decision. Instead, it identifies the
misclassification type from the prior turn feature vector:

\begin{itemize}[leftmargin=*,itemsep=1pt,topsep=2pt]
  \item \textbf{False negative} (unintentional activation executed):
  user hesitates, disengages, cancels, or ignores the system response in case of follow ups.
  \item \textbf{False positive} (intentional activation suppressed):
  user repeats the command, rephrases, or explicitly corrects.
\end{itemize}

This feedback identification step is the critical architectural
distinction. Rather than discarding user reactions, the
system treats them as structured supervision with two error types:
$E_t \in \{\mathrm{FP}, \mathrm{FN}\}$.

We distinguish two labeling processes. The intentional/unintentional activation labels and the
14 acoustic and contextual condition tags used for offline evaluation (\S\ref{sec:experiments})
are human-annotated; they provide the reference ground truth for CAR, UICR, FIR, FAR, and
$P_\text{error}$. The feedback signals used for adaptive pattern induction are not manually
annotated rather they are extracted automatically from the user's next-turn behavior
in multi-turn production logs, for example a same or paraphrased re-invocation within a short
window (false-positive signal), or a cancellation, disengagement, or explicit rejection such as
``stop'' or ``I didn't ask you'' (false-negative signal). Because these behavioral signals can
be ambiguous, we treat them as noisy supervision behavior labels rather than definitive ground-truth labels.

\subsection{ASCIL}
When an automatically inferred feedback signal indicates a potential
misclassification, the system extracts a generalized pattern from the feature
vector $X_t$ of the previous turn and stores it in a structured schema keyed by misclassification
type. For \textit{suppressed\_intentional} errors (false positives),
patterns record \texttt{background\_noise} type, \texttt{word\_count},
\texttt{speech\_energy}, and \texttt{speech\_rate\_wpm}. For
\textit{missed\_unintentional} errors (false negatives), patterns
additionally capture \texttt{preITNText} category (e.g., simple request
vs.\ command) and \texttt{asr\_transcript} clarity which indicates a cross-model transcript
agreement signal between Voxtral Mini 3B's transcript of the audio and the base ASR
transcript. Agreement indicates a reliably-transcribed utterance, while disagreement
signals possible base-ASR rounding to a phonetically nearer valid phrase, a common
false-negative condition. Each pattern includes a human-readable description
injected as context to the LLM at inference. Pattern memory updates as:
\[
P_u^{(t)} = \mathrm{Update}\!\left(P_u^{(t-1)},\, X_t,\, \hat{Y}_t,\, F_t\right)
\]
where $\hat{Y}_t$ is the model's prediction for the current activation and
$F_t$ is an automatically inferred behavioral feedback signal observed after
the interaction. The feedback signal, together with the prior prediction,
determines the feedback-derived error hypothesis used as the pattern key.
The human-annotated reference label $Y_t$ is used only for offline evaluation
and is not available to the online pattern-update process. The model
merges overlapping conditions and discards rare anomalies to prevent
unbounded growth, minimizing cumulative misclassification loss. No raw audio or verbatim transcripts are
retained; only abstract feature signatures are stored for pattern matching and
prompt construction.

\subsection{Deployment}

The correction layer runs in parallel with standard NLU processing. If
an activation is classified as unintentional, the NLU output is silently
suppressed; the user experiences no delay. All pattern storage and
updates occur on-device with no data transmission. At deployment time, the
online pattern-update process requires no additional manual annotation, no
offline retraining, and no modification to base ASR or NLU models, making it a
drop-in layer for any voice assistant stack. Offline evaluation in this work
uses human-annotated reference labels and condition tags, but these labels are
not accessed by the deployed feedback-update mechanism. The correction layer
adds a median latency below 60\,ms on the target hardware in our benchmark with no 
user-perceived latency.

\section{Experiments}
\label{sec:experiments}

\subsection{Dataset}
We evaluate on a proprietary dataset of 3{,}667 interactions from a commercial voice assistant with human-annotated intentional/unintentional reference labels: 2{,}358 intentional activations and 1{,}309 unintentional activations. These reference labels are used only to compute offline evaluation metrics and are distinct from the feedback signals used for adaptive pattern induction, which are inferred automatically from subsequent user behavior. Intentional instances carry multi-label
annotations across 14 overlapping acoustic and contextual conditions:
background noise (stationary, car, non-stationary), background speech
(single person, small group, large group), foreground speech,
pronunciation issues, audio cut-offs, bad audio, multiple speakers, TTS
speech, abandoned utterances, and a clean no-issues baseline.
Table~\ref{tab:categories} summarizes category distribution.

\begin{table}[h]
\centering
\small
\caption{Per-category occurrence counts. Utterances are multi-label;
counts exceed total instances.}
\label{tab:categories}
\setlength{\tabcolsep}{4pt}
\begin{tabular}{lr}
\toprule
\textbf{Category} & \textbf{Count} \\
\midrule
Unintentional wake-up           & 1,309 \\
Background noise -- non-stat.   & 1,593 \\
Background noise -- car         & 483   \\
Background noise -- stationary  & 528   \\
Background speech -- single     & 525   \\
Background speech -- small grp  & 159   \\
Background speech -- large grp  & 6     \\
Pronunciation issue             & 476   \\
Audio cut-off                   & 373   \\
Foreground speech               & 87    \\
Bad audio                       & 100   \\
Multiple speakers               & 3     \\
TTS speech                      & 5     \\
Abandoned utterance             & 8     \\
No issues (clean)               & 100   \\
\bottomrule
\end{tabular}
\end{table}

\subsection{Setup}
\label{sec:setup}
The baseline is a prompt-only classifier built on Qwen3-14B-Instruct,
producing an intent label and confidence score from structured multimodal
prompts without parameter updates, enriched with annotations from Voxtral
Mini 3B, an audio-understanding model. It serves as a
direct ablation of the pattern memory component with same multimodal
features and LLM classifier as ASCIL, but without pattern
storage or feedback-driven correction isolating the adaptive layer's
contribution.

The proposed system augments this baseline with ASCIL.
We use two distinct threshold criteria. For pattern induction, we select baseline misclassifications with confidence $\geq 0.80$ and split them evenly into a feedback-builder subset (N=180) for pattern induction and a held-out subset (N=180) for measuring generalization. This split is session-disjoint. The two subsets share no interaction sessions. We do not enforce a user-disjoint split, since the data spans multi-turn production interactions with substantial per-user variability, and a strict user-disjoint split would leave many users too sparse to evaluate; session-level disjointness preserves evaluation power while avoiding interaction-level overlap between the subsets. Consequently, the held-out subset measures transfer across sessions rather than generalization to unseen users. The human-annotated reference labels are used to identify baseline errors and to compute offline metrics for the builder and held-out subsets. They are not available to ASCIL during online feedback-driven pattern induction or inference. The 0.80 threshold here serves only as a selection criterion for high-confidence errors worth learning from and it is not a reporting threshold. For evaluation, we report all metrics at operating thresholds of 0.85, 0.90, and 0.95. Activations with predicted-unintentional confidence below the threshold are treated as intentional, those at or above as unintentional. Patterns capture SNR ranges, pitch statistics, background noise type, stop-word ratios, POS distributions, and ASR transcript type, with natural language text descriptions to guide the LLM at inference.

\subsection{Metrics}
We report five task-specific metrics: \textbf{CAR} (Correct Acceptance Rate, i.e., the proportion of intentional activations correctly accepted, $\uparrow$); \textbf{UICR} (Unintentional Interception Rate, the proportion of unintentional activations correctly intercepted, $\uparrow$); \textbf{FIR} (False Interception Rate, intentional activations incorrectly suppressed, $\downarrow$); \textbf{FAR} (False Acceptance Rate, unintentional activations incorrectly accepted, $\downarrow$); and \textbf{$P_\text{error}$} (the overall proportion of incorrect decisions, $\downarrow$). All five metrics are computed against the human-annotated reference labels. The system prioritizes precision over recall, suppressing false activations while minimizing the rejection of genuine requests.

\section{Results}
\label{sec:results}

Table~\ref{tab:results_full} presents the complete evaluation across
all conditions and operating thresholds.
\begin{table*}[t]
\centering
\footnotesize
\caption{Complete evaluation results at operating confidence
thresholds (0.85, 0.90, 0.95). CAR: Intentional
Correct Acceptance Rate (\%); UICR: Unintentional Intercept Rate (\%);
FIR: False Interception Rate (\%); FAR: False Acceptance Rate (\%);
$P_\text{error}$: Total Error Rate (\%), computed as
$(\text{FP}{+}\text{FN})/N$. $\Delta P_\text{error}$: relative error
reduction (\%), 
 $(P_\text{error}^\text{base} - P_\text{error}^\text{pattern})/P_\text{error}^\text{base} \times 100$;
positive denotes improvement.
 Per-slice eval counts: No-Issues ($N_\text{int}{=}100$, $N_\text{unint}{=}1{,}309$, total $N{=}1{,}409$); Issue-Tagged ($N_\text{int}{=}2{,}258$, $N_\text{unint}{=}1{,}309$, total $N{=}3{,}567$). The No-Issues slice contains the 100 intentional utterances tagged as no-issues (clean audio); the Issue-Tagged slice contains the remaining 2{,}258 intentional utterances (each carrying at least one acoustic or contextual condition tag). The 1{,}309 unintentional samples are shared across both slices, so UICR and FAR are identical by construction between the No-Issues and Issue-Tagged conditions at a given threshold; only CAR and FIR vary. Total intentional accounting: 100 (No-Issues) + 2{,}258 (Issue-Tagged) = 2{,}358. The Held-Out set (N=180) is session-disjoint from the feedback-builder set (N=180) used for pattern induction and is the only split intended to measure transfer beyond the feedback-builder interactions; it is session-disjoint rather than user-disjoint. The Issue-Tagged slice includes builder-derived cases and therefore measures overall system behavior rather than clean held-out generalization.}
\label{tab:results_full}
\setlength{\tabcolsep}{4pt}
\resizebox{\textwidth}{!}{%
\begin{tabular}{llccccccc}
\toprule
\textbf{Condition} & \textbf{System} & \textbf{Thresh.} &
\textbf{CAR}$\uparrow$ & \textbf{UICR}$\uparrow$ & \textbf{FIR}$\downarrow$ &
\textbf{FAR}$\downarrow$ & \textbf{$P_\text{error}$}$\downarrow$ &
\textbf{$\Delta P_\text{error}$} \\
\midrule
\multirow{6}{*}{\makecell[l]{No-Issues\\(clean Intentional audio)}}
& Baseline (only model)  & 0.85 & 94.00 & \textbf{55.31} & 6.00 & \textbf{44.69} & \textbf{41.94} & --- \\
& model with ASCIL layer & 0.85 & \textbf{98.00} & 43.16 & \textbf{2.00} & 56.84 & 52.95 & \textcolor{lossred}{$-$26.25} \\
\cmidrule{2-9}
& Baseline (only model)    & 0.90 & 97.00 & 25.44 & 3.00 & 74.56 & 69.48 & --- \\
& model with ASCIL layer   & 0.90 & \textbf{98.00} & \textbf{40.72} & \textbf{2.00} & \textbf{59.28} & \textbf{55.22} & \textcolor{gaingreen}{\textbf{+20.52}} \\
\cmidrule{2-9}
& Baseline (only model)    & 0.95 & 100.00 & 2.83 & 0.00 & 97.17 & 90.28 & --- \\
& model with ASCIL layer   & 0.95 & 100.00 & \textbf{17.95} & 0.00 & \textbf{82.05} & \textbf{76.22} & \textcolor{gaingreen}{\textbf{+15.57}} \\
\midrule
\multirow{6}{*}{
  \makecell[l]{
    Issue-Tagged\\
    (multi-condition intentional\\
    activations resembling\\
    unintentional activations)
  }
}
& Baseline (only model)   & 0.85 & 79.05 & \textbf{55.31} & 20.95 & \textbf{44.69} & 29.66 & --- \\
& model with ASCIL layer   & 0.85 & \textbf{93.18} & 43.16 & \textbf{6.82} & 56.84 & \textbf{25.18} & \textcolor{gaingreen}{\textbf{+15.10}} \\
\cmidrule{2-9}
& Baseline (only model)   & 0.90 & 89.15 & 25.44 & 10.85 & 74.56 & 34.23 & --- \\
& model with ASCIL layer   & 0.90 & \textbf{93.49} & \textbf{40.72} & \textbf{6.51} & \textbf{59.28} & \textbf{25.88} & \textcolor{gaingreen}{\textbf{+24.39}} \\
\cmidrule{2-9}
& Baseline (only model)   & 0.95 & \textbf{99.78} & 2.83 & \textbf{0.22} & 97.17 & 35.80 & --- \\
& model with ASCIL layer   & 0.95 & 97.21 & \textbf{17.95} & 2.79 & \textbf{82.05} & \textbf{31.88} & \textcolor{gaingreen}{\textbf{+10.94}} \\
\midrule
\multirow{2}{*}{\makecell[l]{Held-Out\\(all failing cases)}}
& Baseline (only model) & 0.90 & \multicolumn{5}{c}{\textit{consistent failure on all samples ($P_\text{error}{=}100\%$ by construction)}} & --- \\
& model with ASCIL layer   & 0.90 & \textbf{70.32} & \textbf{39.88} & \textbf{29.68} & \textbf{60.12} & \textbf{45.73} & \textcolor{gaingreen}{\textbf{+54.27}} \\
\bottomrule
\end{tabular}%
}
\end{table*}

\subsection{Session-disjoint held-out subset}
The held-out subset comprises interactions misclassified by the baseline
across both false-positive and false-negative types. So baseline
$P_{\text{error}}$ is 100\% on this subset by construction (UICR\,=\,0\%
on all unintentional cases). The pattern layer recovers meaningful
performance, achieving 70.32\% CAR and 39.88\% UICR at threshold 0.90
and reducing $P_{\text{error}}$ to 45.73\%; this corresponds to a
\textbf{54.27\% relative error reduction} against the failure
baseline. Because this subset is constructed entirely from baseline failures,
the figure reports conditional recovery from those failures rather than an
absolute error reduction over the full dataset. This suggests that the learned
patterns transfer to session-disjoint, structurally similar misclassification
cases. Because the split is session-disjoint but not user-disjoint, this result
measures transfer across sessions for users who may occur in both subsets,
rather than generalization to unseen users.

\subsection{Intenional activations with multi labeled Issues-tagged evaluation slice.}
On the intentional with issue-tagged evaluation slice (2{,}258 intentional utterances carrying
at least one issue condition tag, plus the shared 1{,}309 unintentional utterances),
ASCIL reduces $P_\text{error}$ at all reported thresholds, with the largest
relative reduction of \textbf{24.39\%} at threshold 0.90. This slice includes
builder-derived cases and is therefore a system-level diagnostic rather than a
contamination-free held-out test. The issue-tagged intentional activations combine multiple acoustic and contextual conditions, making them closely resemble unintentional activations. These challenging cases involve
simultaneous multi-factor conditions like background noise, foreground
speech, and pronunciation variation occurring together. At threshold 0.90, ASCIL improves both intentional acceptance and
unintentional interception: CAR rises from 89.15\% to 93.49\%, while UICR
improves from 25.44\% to 40.72\%. Correspondingly, FIR decreases from
10.85\% to 6.51\% and FAR decreases from 74.56\% to 59.28\%. These
simultaneous improvements are threshold-dependent. At other thresholds, ASCIL
exhibits different acceptance--interception trade-offs.

\subsection{No-issues dataset.}
On clean audio, the ASCIL produces mixed results at threshold 0.85 ($-26.25\%$ $\Delta P_\text{error}$). At this lower threshold, the pattern layer shifts decisions toward acceptance. FIR improves (6.00\% $\rightarrow$ 2.00\%) and CAR rises (94.00\% $\rightarrow$ 98.00\%), but FAR worsens (44.69\% $\rightarrow$ 56.84\%) and UICR falls (55.31\% $\rightarrow$ 43.16\%). Because the unintentional class dominates the No-Issues evaluation slice (1{,}309 vs. 100 samples), the FAR increase drives the overall rise in $P_\text{error}$. This behavior reflects a limitation at the 0.85 threshold and motivates further evaluation of operating points at or above 0.90. At operating thresholds of 0.90 and 0.95, the
ASCIL recovers strongly, reducing $P_\text{error}$ by 20.52\%
and 15.57\% respectively. At threshold 0.90, CAR improves from 97.00\% to 98.00\% and UICR rises substantially from 25.44\% to 40.72\%. At threshold 0.95, UICR improves from 2.83\% to 17.95\% while CAR remains at 100\%. This precision-first behavior of filtering edge-case false
positives without degrading strong true positives is particularly
valuable in clean-audio conditions where false suppression of genuine
requests directly erodes user trust. Because the unintentional samples are shared across
the No-Issues and Issue-Tagged slices, UICR and FAR are identical by construction between
the two conditions. The CAR and FIR differences are where the discriminative signal lies.
Issue-Tagged intentional utterances carry acoustic and contextual conditions (background
noise, background speech, pronunciation issues, cutoffs) that acoustically resemble
unintentional wake-ups, making them harder to classify correctly. The CAR and FIR
improvements observed on Issue-Tagged therefore reflect improved discrimination on this
harder population, not a change in unintentional detection.

\subsection{Interpretation.}
The results reveal a consistent pattern across conditions. The adaptive
layer is most effective at the higher operating thresholds (0.90--0.95)
where the model is operating at high confidence. In this
framework, the ASCIL adds targeted corrections for the specific
acoustic and contextual signatures it has learned, without disrupting
the base model's already-reliable judgments. From a user-experience
perspective, this translates to fewer disruptive unintended responses
and fewer suppressed genuine commands. These are the two failure modes that
most directly erode trust in voice interfaces.
\section{Conclusion}
\label{sec:conclusion}
We introduced a feedback-driven ASCIL that closes a specific gap left open by
prior post-ASR  approaches. A closed loop that learns from
user reactions \emph{after} a wake-up has already been handled by treating implicit and explicit user feedback as
noisy behavioral supervision, extracting generalized on-device patterns, and
applying corrections at inference without retraining. This framework
achieves meaningful, consistent improvements on real-world multi-condition
audio. The layer autonomously decides when to extract patterns from feedback signals and when to apply them at inference, without external orchestration. The current architecture keeps added
latency low by running in parallel to NLU and suppressing the response rather than the request, generalizing beyond false wake-up suppression to any conversational
AI setting where systematic misclassifications are exposed by user feedback.
We present these results as a preliminary systems study rather than definitive
evidence of personalized generalization or production-level efficacy.

\section*{Limitations}
ASCIL's effectiveness depends on sufficient user interaction history, leaving
cold-start adaptation for new users as a known challenge. Feedback signals are
ambiguous like silence for example may indicate disengagement, satisfaction, or a
natural pause. The current study does not independently validate the precision
and recall of the feedback heuristics. The evaluation uses a  proprietary
dataset with session-disjoint rather than user-disjoint splits, so
generalization across unseen users, languages, device categories, and
interaction styles remains a part of future work.

\section*{Ethics Statement}
Interactions were collected from users who consented to data usage for product
improvement under the product's terms of service. Samples were de-identified,
with personally identifiable information removed from content prior to annotation.
Users are referenced only by pseudonymous stable identifiers that support
session-level grouping without identity linkage. Annotators accessed data only
through a controlled labeling interface, and data retention follows standard
company policy. Demographics are not tracked at the interaction level for the
evaluation dataset used in this work.

\bibliography{refs}

\end{document}